\documentclass[letterpaper, 10 pt, journal, twoside]{ieeetran}  % Comment this line out if you need a4paper

\IEEEoverridecommandlockouts                              % This command is only needed if 
\usepackage{amsmath,amsfonts}
\usepackage{amssymb}

\usepackage[ruled, lined, longend, linesnumbered]{algorithm2e}
\usepackage{color}

\usepackage{array}
\usepackage{textcomp}
\usepackage{stfloats}

\usepackage{enumitem}

\usepackage{verbatim}
\usepackage{graphicx}
\usepackage{threeparttable}

\usepackage{cite}
\usepackage[utf8]{inputenc} % allow utf-8 input
\usepackage[T1]{fontenc}    % use 8-bit T1 fonts
\usepackage{hyperref}       % hyperlinks
\usepackage{url}            % simple URL typesetting
\usepackage{booktabs}       % professional-quality tables
\usepackage{amsfonts}       % blackboard math symbols
\usepackage{nicefrac}       % compact symbols for 1/2, etc.
\usepackage{microtype}      % microtypography
\usepackage{lipsum}
\usepackage{fancyhdr}       % header
\allowdisplaybreaks

\usepackage{graphicx}       % graphics
\usepackage{wrapfig}
\usepackage[export]{adjustbox}
\usepackage[position=top]{subfig}
\usepackage{subcaption} % Use subcaption for subfigures and subtables
\usepackage{mwe}

\usepackage{multirow}
\usepackage{makecell}
\usepackage{balance}

\usepackage{tcolorbox}

\title{
DISCO: Language-Guided Manipulation with Diffusion Policies \\ and Constrained Inpainting
}

\author{Ce Hao$^{*1}$, Kelvin Lin$^{1}$, Zhiwei Xue$^{1}$, Siyuan Luo$^{1}$, Harold Soh$^{*1, 2}$
\thanks{
This research is supported by A*STAR under its National Robotics Programme (NRP) (Award M23NBK0053). 
} %Use only for final RAL version
\thanks{$^{1}$ School of Computing, National University of Singapore.}
\thanks{$^2$ Smart Systems Institute, NUS.}
\thanks{$^*$correspondence to {\tt\footnotesize cehao@u.nus.edu} and {\tt\footnotesize harold@nus.edu.sg}}

}
\newcommand{\rev}[1]{\textcolor{black}{#1}}

\newcommand{\revcolorbegin}{\color{black}}
\newcommand{\revcolorend}{\color{black}}

\begin{document}

\maketitle
%\thispagestyle{empty}
%\pagestyle{empty}

%%%%%%%%%%%%%%%%%%%%%%%%%%%%%%%%%%%%%%%%%%%%%%%%%%%%%%%%%%%%%%%%%%%%%%%%%%%%%%%%

\begin{abstract}
Diffusion policies have demonstrated strong performance in generative modeling, making them promising for robotic manipulation guided by natural language instructions. However, generalizing language-conditioned diffusion policies to open-vocabulary instructions in everyday scenarios remains challenging due to the scarcity and cost of robot demonstration datasets. To address this, we propose DISCO, a framework that leverages off-the-shelf vision-language models (VLMs) to bridge natural language understanding with high-performance diffusion policies.
DISCO translates linguistic task descriptions into actionable 3D keyframes using VLMs, which then guide the diffusion process through constrained inpainting. However, enforcing strict adherence to these keyframes can degrade performance when the VLM-generated keyframes are inaccurate. To mitigate this, we introduce an inpainting optimization strategy that balances keyframe adherence with learned motion priors from training data. Experimental results in both simulated and real-world settings demonstrate that DISCO outperforms conventional fine-tuned language-conditioned policies, achieving superior generalization in zero-shot, open-vocabulary manipulation tasks.
Videos see website: 
\href{https://sites.google.com/view/disco2025/}{sites.google.com/view/disco2025}.
\end{abstract}

% TODO：

\section{Introduction} \label{Sec: intro}
\IEEEPARstart{R}{obotic} manipulation remains a fundamental challenge due to the complexity of handling diverse objects across varying contexts. Recent advancements in diffusion-based policies have significantly improved robot capabilities, enabling the successful execution of multi-modal, long-horizon, and multi-object tasks
~\cite{chi2023diffusion, ze20243d, janner2022planning, hao2025chd, bi2025vla, yu2024manipose, yu2024uniaff, li2024skt, yu2025generalizable, yu2024gamma}.
A key next step is language-conditioned control of diffusion policies, allowing robots to follow natural language instructions for more flexible and adaptable behavior
~\cite{zhou2023language, hao2025chd}
. By integrating language as a control mechanism, an otherwise unconditional diffusion policy can be adapted to specific task contexts, paving the way for more generalizable and interactive robot manipulation.

Existing approaches directly input language instructions into diffusion policies to guide action generation
~\cite{stepputtis2020language, reuss2023goal, ha2023scaling}. 
However, fine-tuned language-conditioned policies often struggle with unseen tasks and open-vocabulary instructions~\cite{stepputtis2020language, liu2024moka, brohan2022rt}. This limitation becomes evident in tasks like object grasping, where even policies trained on large multi-modal datasets frequently fail to execute correct grasps when faced with novel natural language instructions
~\cite{shridhar2023perceiver, lin2023text2motion, yu2024manipose}.

\begin{figure}[t]
    \centering
    \includegraphics[width=0.9\columnwidth]{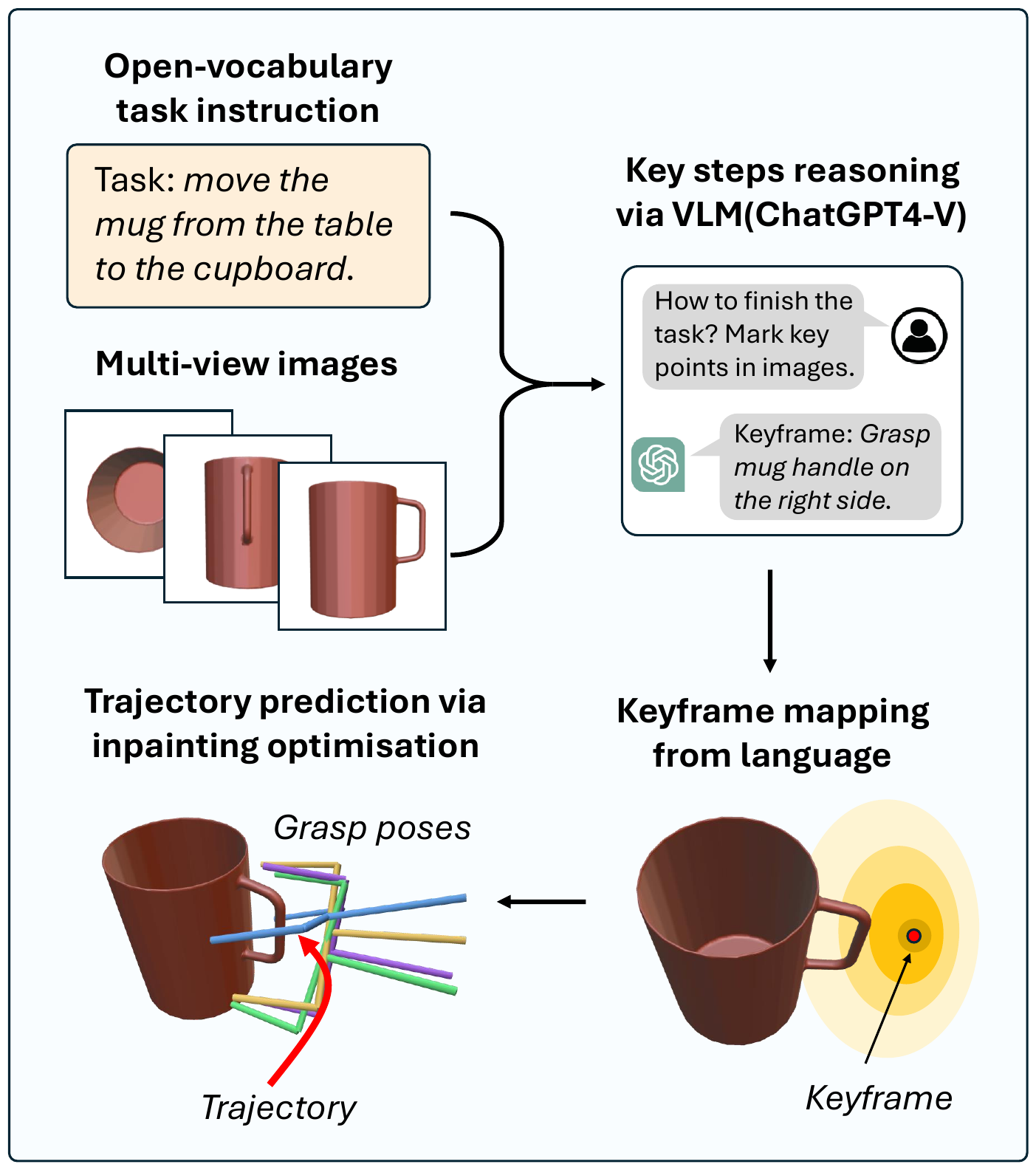}
    \caption{An overview of the DISCO framework, which interprets task descriptions and multi-view observations to generate keyframes in 3D space, which are used to guide the diffusion-based generation of actions (e.g., grasp poses) through constrained inpainting optimization. }
    \label{Fig: teaser}
\vspace{-7mm}
\end{figure}

In this paper, we propose a framework to integrate general-purpose ``off-the-shelf'' vision-language models (VLMs) and diffusion policy for open-vocabulary robot manipulation, which we call \textbf{DISCO} (\textbf{D}iffusion \textbf{I}npainting with \textbf{S}emantic Keyframes and \textbf{C}onstrained \textbf{O}ptimization) 
As in Figure.~\ref{Fig: teaser}, our key idea is that by integrating VLM-generated \textit{keyframes} as actionable guides in a diffusion process, we can bridge the gap between high-level linguistic instructions and low-level robot actions. Different from prior work~\cite{shridhar2023perceiver}, the keyframes are \emph{not} precise subgoals but serve as ``coarse'' guidance. 
Here, we use a general VLM model (e.g., ChatGPT4) to convert task descriptions into keyframes in 3D space.
Then, we employ an inpainting method that uses these keyframes to guide the policy's action sequence generation. Since the keyframes serve only as rough guides, standard inpainting~\cite{lugmayr2022repaint, rout2023theoretical} --- which enforces adherence to the keyframes --- can fail to generate successful trajectories. This occurs since keyframes can be inaccurate or far outside the support of the training distribution~\cite{yang2023compositional}. To further enhance robustness, we propose an inpainting optimization method that constrains the generated points to lie within regions of sufficient support during diffusion. This approach balances keyframe adherence and remaining within the support of the diffusion model, which results in better performance for novel or ambiguous task descriptions.

We validate our open-vocabulary diffusion policy, DISCO, using simulated benchmarks~\cite{sharma2018multiple, urain2023se} and a real-robot grasping testbed~\cite{chen2024behavioral}. The simulation experiments include adapted language-guided block pushing, object grasping, and long-horizon tasks such as the Franka Kitchen and multi-task Calvin benchmark. These tasks require generating continuous actions, including end-effector positions, velocities, and grasp poses. 
Compared to fine-tuned diffusion policy baselines, DISCO achieves comparable performance on seen tasks while significantly outperforming them on tasks with novel language instructions. Additionally, we conduct an ablation study comparing the use of curves generated by vision-language models (VLMs) as intermediate representations with the keyframes used in DISCO~\cite{gu2023rt}. Finally, we evaluate zero-shot transfer to real-world language-guided grasping, where DISCO successfully executes grasping in real-time, manipulating various objects under human instructions without fine-tuning.

\noindent In summary, the contribution of this paper is threefold:
\begin{enumerate}[noitemsep, leftmargin=10pt]
    \item We proposed the DISCO framework that integrates off-the-shelf VLMs to diffusion policy via keyframes for language-guided robot manipulation.
    \item We designed the inpainting optimization method to circumvent inaccurate and out-of-distribution keyframes generated by open-vocabulary instructions.
    \item We conducted extensive experiments in both simulation and real-world robots, showcasing DISCO's performance on continuous action and grasping pose generation over baselines. 
\end{enumerate}

\section{Related Works} \label{Sec: related}

\textbf{Language-conditioned robot manipulation}, where natural language is used to guide robot behavior, has been receiving increased attention, partially due to the advances made in large-language models. One line of prior work has applied vision-language models (VLMs) to convert linguistic descriptions into executable trajectories
~\cite{zhou2023language, stepputtis2020language, driess2023palm, huang2023voxposer}
; however, direct trajectory generation with general-purpose VLMs often results in unreliable or infeasible motions, especially in open-vocabulary or unstructured settings. Other methods use VLMs to transform language descriptions into code, affordances, or actionable trajectory tokens~\cite{brohan2022rt, brohan2023rt, shridhar2023perceiver, liang2023code}, but affordance approaches typically require task-specific fine-tuning and do not naturally generalize to multi-view 3D reasoning, while code-generation methods do not provide interpretable intermediate signals for action guidance. To address these challenges, our method introduces keyframes as a more reliable and generalizable intermediate representation to bridge high-level language commands and low-level trajectories.

\textbf{Diffusion-based policies} are a rich policy class that can represent multi-modal and high-dimensional action distributions. They have been shown to be excellent for imitation learning, where they can be trained to exhibit intricate and complex motion
~\cite{chi2023diffusion, janner2022planning, li2023crossway, chen2024behavioral, ze20243d, xian2023chaineddiffuser}
. Recent advancements have expanded diffusion policy applications to include goal-conditioned and language-conditioned tasks~\cite{ho2022classifier, reuss2023goal, ha2023scaling}, providing a structured approach for integrating explicit objectives and linguistic instructions into action generation~\cite{zhou2023language, zhang2022language}. However, these methods typically require extensive fine-tuning for each specific task, limiting their ability to generalize to unseen, open-vocabulary tasks
~\cite{ahn2022can, zhi2024closed}
.
To tackle this, we propose a zero-shot transfer strategy using inpainting, complemented by constrained optimization to handle out-of-distribution keyframes resulting from novel task descriptions.

\textbf{Inpainting} was originally developed for image completion~\cite{elharrouss2020image}, but has been adapted for use with diffusion policies to generate actions rather than images~\cite{ho2020denoising, lugmayr2022repaint}. These methods have also been applied to goal-conditioned planning, using goals as known conditions to guide action generation~\cite{yang2023planning, rout2023theoretical, janner2022planning}. In our approach, we apply inpainting within the diffusion policy framework to generate trajectories and grasp poses, employing keyframes as guidance for action sequences.

\section{Preliminaries} \label{Sec: prelim}

\subsection{Problem Formulation}
We define the language-conditioned manipulation task as controlling the robot to manipulate objects to achieve a target state under a language specification (or task description) $l$.
The language specifications are ``open-vocabulary'' in that they contain words and phrases in natural language, beyond a small predefined set of labels. The behavior of the robot should comply with the specification. In this work, we have access to image observations $O$ of the robot and objects. The set of observations is assumed to be sufficient to complete the task; for example, multiple views may be available to provide sufficient state information. We seek a policy $\pi$ that, at each time step $t$, generates an action (sequence) $\boldsymbol{a}$ conditioned on the current observation $O_t$ and language description $l$, i.e.,  $\boldsymbol{a}_t \sim \pi(\boldsymbol{a} \mid O_t, l)$. The task is deemed successful when the target is reached and the robot trajectories satisfy the language specifications. In the following, we will drop the explicit dependence on the time step to avoid clutter. 

\subsection{Diffusion Policy}
Diffusion Policy~\cite{chi2023diffusion, ke20243d, ha2023scaling} applies diffusion-based generative models towards robot action generation. In brief, the model assumes that   $p_\theta\left(\boldsymbol{x}_0\right):=\int p_\theta\left(\boldsymbol{x}_{0: N}\right) d \boldsymbol{x}_{1: N}$, where $\boldsymbol{x}_1, \ldots, \boldsymbol{x}_N$ are latent variables of the same dimensionality as the data $\boldsymbol{x}_0 \sim p\left(\boldsymbol{x}_0\right)$. A forward diffusion chain gradually adds noise to the data $\boldsymbol{x}_0 \sim q\left(\boldsymbol{x}_0\right)$ in $N$ steps with a pre-defined variance schedule $\beta_i$, as,
\vspace{-1mm}
\begin{align}
    & q\left(\boldsymbol{x}_{1: N} \mid \boldsymbol{x}_0\right):=\prod_{t=1}^N q\left(\boldsymbol{x}_i \mid \boldsymbol{x}_{i-1}\right), \\
    & q\left(\boldsymbol{x}_i \mid \boldsymbol{x}_{i-1}\right):=\mathcal{N}\left(\boldsymbol{x}_i ; \sqrt{1-\beta_i} \boldsymbol{x}_{i-1}, \beta_i \boldsymbol{I}\right).
\end{align}

In DDPM~\cite{ho2020denoising}, the reverse process is modeled as a Gaussian distribution, and the reverse diffusion chain is constructed as,
\vspace{-1mm}
\begin{align}
    & p_\theta\left(\boldsymbol{x}_{0: N}\right):=\mathcal{N}\left(\boldsymbol{x}_N ; \boldsymbol{0}, \boldsymbol{I}\right) \prod_{t=1}^N p_\theta\left(\boldsymbol{x}_{i-1} \mid \boldsymbol{x}_i\right) \\
    & p_\theta\left(\boldsymbol{x}_{i-1} \mid \boldsymbol{x}_i\right)=\mathcal{N}\left(\boldsymbol{x}_{i-1} ; \mu_\theta\left(\boldsymbol{x}_i, i\right), \Sigma_\theta\left(\boldsymbol{x}_i, i\right)\right),
\end{align}
which is then optimized by maximizing the evidence lower bound defined as $\mathbb{E}_q\left[\ln \frac{p_\theta\left(\boldsymbol{x}_{0: N}\right)}{q\left(\boldsymbol{x}_{1: N} \mid \boldsymbol{x}_0\right)}\right]$. After training, sampling from the diffusion model consists of sampling $\boldsymbol{x}_N \sim p\left(\boldsymbol{x}_N\right)$ and running the reverse diffusion chain to go from $i=N$ to $0$. The diffusion policy is formulated as the reverse process of a conditional diffusion model to generate actions,
\vspace{-1mm}
\begin{equation}
\begin{aligned}
    \pi_\theta(\boldsymbol{a} \mid O) & =  p_\theta\left(\boldsymbol{a}_{0: N} \mid O\right) \\
    & =\mathcal{N}\left(\boldsymbol{a}_N ; 0, I\right) \prod_{i=1}^N p_\theta\left(\boldsymbol{a}_{i-1} \mid \boldsymbol{a}_i, O\right), 
\end{aligned} \label{Eqn: diffusion policy formulation}
\end{equation}
\noindent where the end sample of the reverse chain, $\boldsymbol{a}_0$, is the action used for execution.
\vspace{-1mm}
\subsection{Inpainting}
Inpainting~\cite{liu2023image, elharrouss2020image, janner2022planning} was originally formulated for image generation, where part of an image is known $\boldsymbol{x}^{\text{known}}$ and the other part is ``masked'' (or unknown) $\boldsymbol{x}^{\text{unknown}}$. Standard inpainting  integrates the $\boldsymbol{x}^{\text{known}}$ and $\boldsymbol{x}^{\text{unknown}}$ during the reverse process to reconstruct the missing regions~\cite{lugmayr2022repaint}, 
\begin{align}
&  \boldsymbol{x}_{i - 1}^{\text{known}}   \sim p_{i-1}^{\text{known}} = \mathcal{N}\left( {\sqrt{{\bar{\alpha}}_{i}}\boldsymbol{x}_{0}^{\text{known}},\left( {1 - \bar{\alpha}_i} \right)I} \right), \label{Eqn: inpt_known} \\
&  \boldsymbol{x}_{i - 1}^{\text{unknown}}   \sim q_{i-1}^{\text{unknown}} = \mathcal{N}\left( {\mu_{\theta}\left( {\boldsymbol{x}_{i},i} \right),\Sigma_{\theta}\left( {\boldsymbol{x}_{i},i} \right)} \right), \label{Eqn: inpt_unknown} \\
& \boldsymbol{x}_{i - 1}  = \boldsymbol{m} \odot \boldsymbol{x}_{i - 1}^{\text{known}} + (1 - \boldsymbol{m}) \odot \boldsymbol{x}_{i - 1}^{\text{unknown}}. \label{Eqn: vanilla inpainting}
\end{align} 

In Eqn.~\eqref{Eqn: inpt_known} and \eqref{Eqn: inpt_unknown}, $\boldsymbol{x}_{i - 1}^{\text{known}}$ and $\boldsymbol{x}_{i - 1}^{\text{unknown}}$ denote the known part of the image and the unknown inpainting parts, sampled from the probabilities $p_{i - 1}^{\text{known}}$ and $q_{i - 1}^{\text{unknown}}$ at  $(i-1)_{\text{th}}$ step.  $\bar{\alpha}_i=$ $\prod_{s=1}^i\left(1-\beta_s\right)$, where $\beta_s$ controls the variance of the Gaussian noise added at each forward step. $\mu_{\theta}$ and $\Sigma_{\theta}$ represent the scheduling parameters and parameters of the learned diffusion distribution~\cite{lugmayr2022repaint, ho2020denoising}. 
$\boldsymbol{m}$ represents the inpainting mask, which takes values of 1 for known regions and 0 for unknown regions; $\odot$ denotes the element-wise production.
During the reverse diffusion, $\boldsymbol{x}_{i - 1}$ is constructed by merging  $\boldsymbol{x}_{i - 1}^{\text{known}}$ and $\boldsymbol{x}_{i - 1}^{\text{unknown}}$. This integration ensures the content generated for the unknown regions is conditioned on the known regions, leading to a coherent output.

\section{Method: Diffusion Inpainting with Semantic Keyframes and Constrained Optimization} \label{Sec: method}

\begin{figure*}
    \centering
    \includegraphics[width=0.9\textwidth]{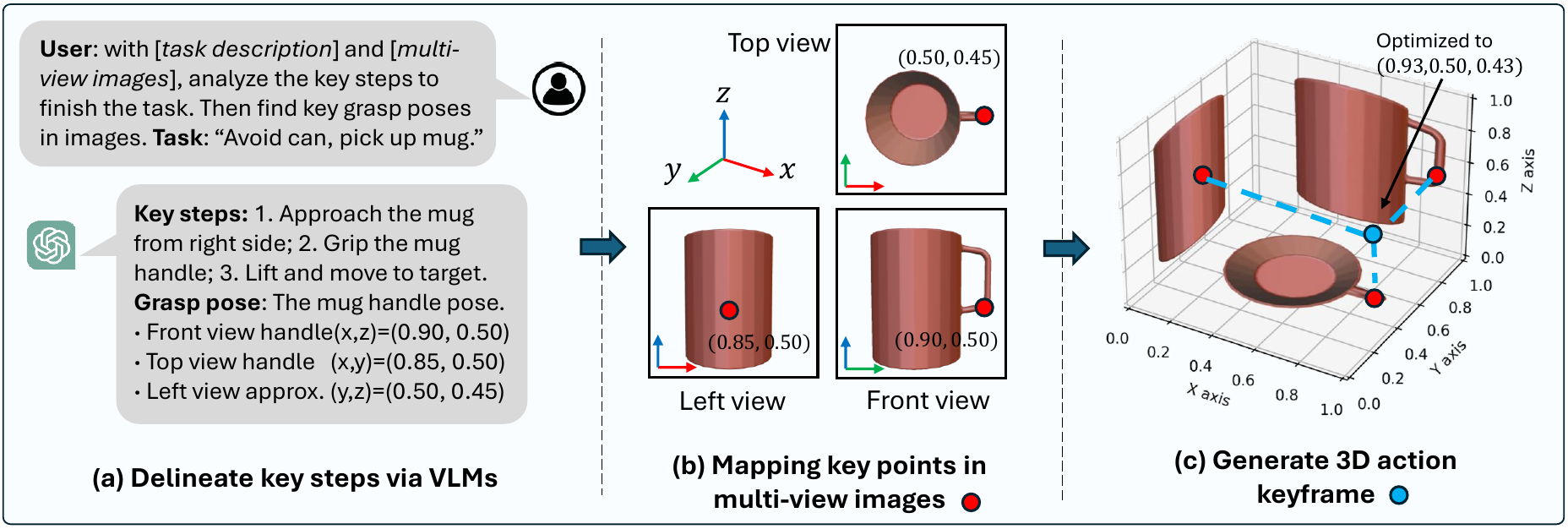}
    \caption{DISCO predicts keyframes by (a) delineating the key steps to finish the task; (b) mapping key point positions in images; and (c) generating the 3D action keyframe based on multi-view key points.}
    \label{Fig: keyframe}
\vspace{-5mm}
\end{figure*}

This section describes DISCO, our contribution towards zero-shot language-conditioned diffusion.
We introduce the keyframe generation method (Sec.~\ref{Subsec: keyframe gen}) followed by the constrained inpainting optimization (Sec.~\ref{Subsec: vanilla inpainting}) to address issues that arise due to poor or invalid keyframes generated by open-vocabulary descriptions. 
DISCO is summarized in \textbf{Algorithm~\ref{Alg: disco}}.

\begin{figure}
    \centering
    \includegraphics[width=0.97\columnwidth]{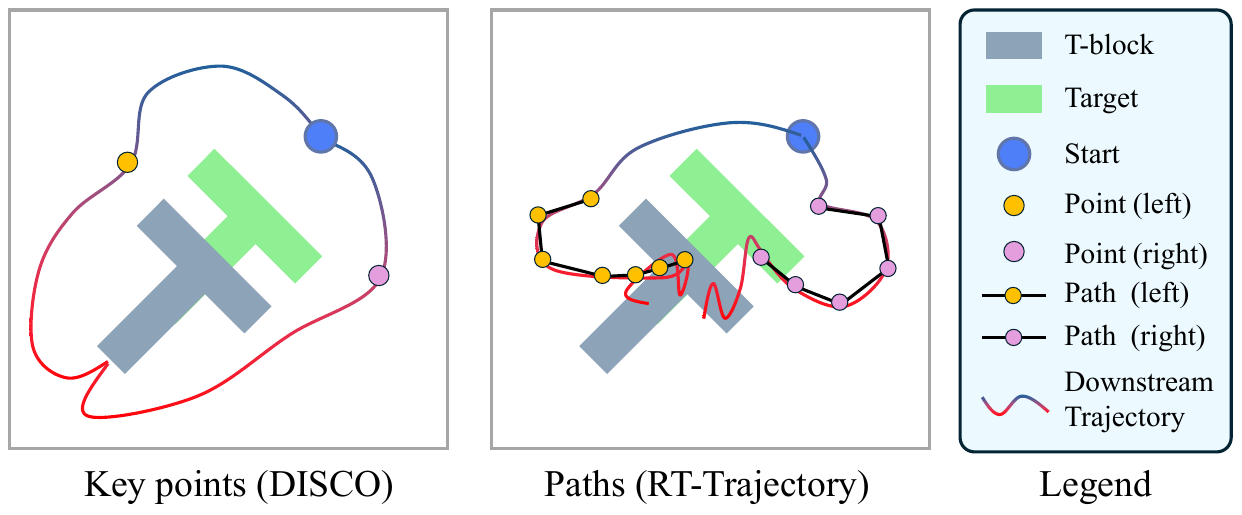}
    \caption{\rev{Comparison of VLM-generated points versus paths on the Push-T task. The agent is instructed to push the block from the left or right side. The figures depict the generated points/paths by ChatGPT4o and the resulting trajectories using DISCO (for points) and RT-Trajectory~\cite{gu2023rt} (for paths). We found ChatGPT4o generated key points more reliably than paths.} }
    \label{Fig: RT-Traj}
\vspace{-6mm}
\end{figure}

\subsection{Keyframe Generation from Language Descriptions} \label{Subsec: keyframe gen}
As illustrated in Figure~\ref{Fig: keyframe}, we employ general multi-modal vision-language models (VLMs), such as ChatGPT4, to analyze the task description $l$ and observation $O$, and generate action keyframes using three main steps: 
\begin{enumerate}
\item \textbf{Delineating the key steps} necessary to complete the task~\cite{huang2023voxposer, driess2023palm}. 
For example, given the task ``Avoid the can and pick up the mug'', the VLM generates two key steps: ``First move to the mug from the right side of the can'' then ``Grasp the mug handle and lift it.'' 

\item \textbf{Mapping key points} $\boldsymbol{p}^{\text{key}}$ on the observations $O$. 
We ask the VLMs to map corresponding key points $\boldsymbol{p}^{\text{key}}$ based on the above steps by marking the relevant points on the images $O$. For example, grasping the mug handle would involve marking the handle region. We use points rather than paths because preliminary tests indicated that points were generated more robustly by current off-the-shelf VLMs (see illustrative example in Fig. 3). DISCO can potentially be used on paths, which would be interesting future work. Moreover, note that the points marked by VLMs are not exact grasp points (unlike  \cite{shridhar2022cliport, shridhar2023perceiver, li2023manipllm}) and are not used to directly plan manipulation trajectories. Instead, they serve to provide ``coarse'' information that guides trajectory generation. 

\item \textbf{Generating action keyframes} $\boldsymbol{a}^{\text{key}}$ in the environment. Given the key points of the end-effector, we convert them to the keyframes $\boldsymbol{a}^{\text{key}}$ based on the policy and environment. \rev{Depending on the exact application, $\boldsymbol{a}^{\text{key}}$ could represent the 3D position, or be converted to end-effector velocity or joint configuration. In practice, we first calibrate a mapping from image keypoints to 3D positions, then use the robot's current state and inverse kinematics to obtain the actionable keyframe.
}
\end{enumerate}

\begin{figure*}[t]
    \centering
    \includegraphics[width=0.8\textwidth]{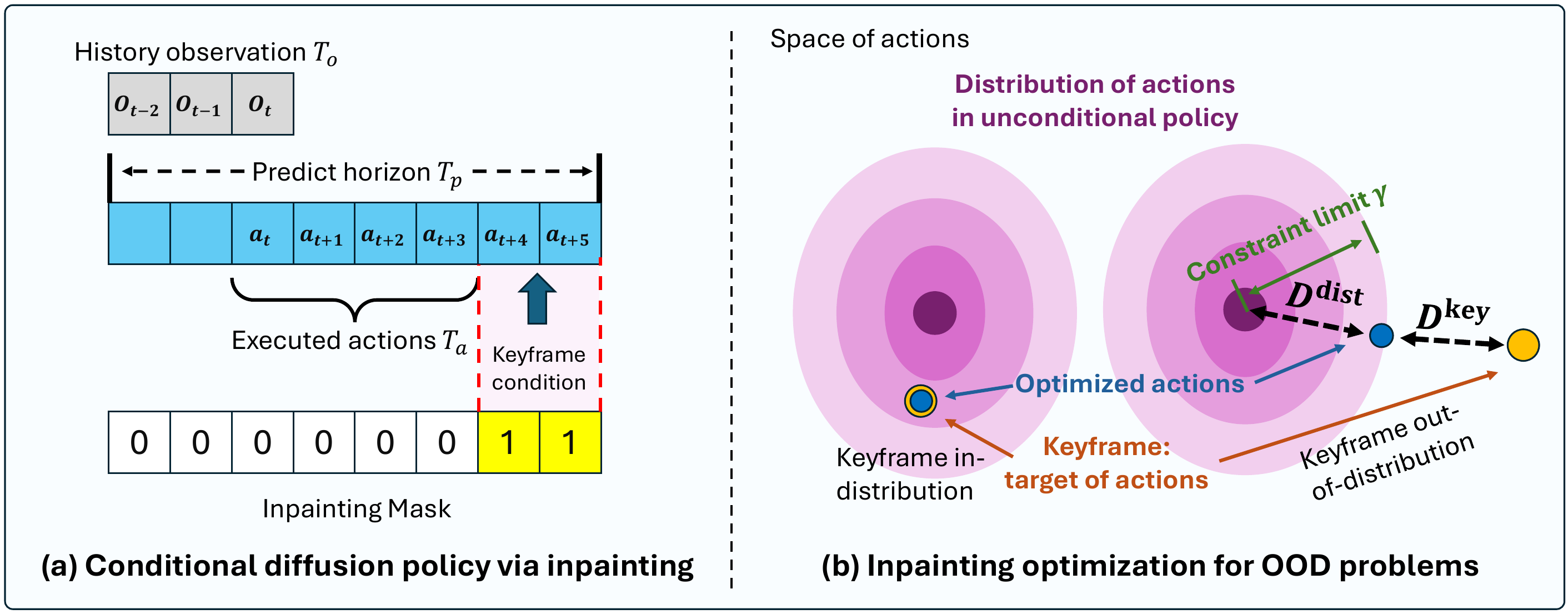}
    \caption{\textbf{(a)}: Inpainting controls the trajectory generation of diffusion policy via keyframes. \textbf{(b)}: Our inpainting optimization method constrains the generated actions to lie within regions of sufficient support (high likelihood). $D^{\text{key}}$ denotes the distance of masked action to keyframe in Eqn.~\eqref{Eqn: inpt opt obj}; $D^{\text{dist}}$ represents the negative log-likelihood constraint in Eqn.~\eqref{Eqn: inpt opt const}. }
    \label{Fig: inpainting}
\vspace{-5mm}
\end{figure*}

\subsection{Conditional Diffusion Policy via Inpainting} \label{Subsec: vanilla inpainting}

After generating the keyframes $\boldsymbol{a}^{\text{key}}$, we employ inpainting to perform keyframe-conditioned diffusion-based action generation $\boldsymbol{a} \sim \pi(\boldsymbol{a} | O, \boldsymbol{a}^{\text{key}}) = \mathcal{N}\left(\boldsymbol{a}_N ; 0, I\right) \prod_{i=1}^N p_\theta\left(\boldsymbol{a}_{i-1} \mid \boldsymbol{a}_i, O, \boldsymbol{a}^{\text{key}}\right)$  (from Eqn.~\eqref{Eqn: diffusion policy formulation}), where inpainting of the conditional diffusion model is formulated as,
\vspace{-1mm}
\begin{align}
& \boldsymbol{a}_{i - 1}  = \boldsymbol{m} \odot \boldsymbol{a}_{i - 1}^{\text{known}} + (1 - \boldsymbol{m}) \odot \boldsymbol{a}_{i - 1}^{\text{unknown}} \label{Eqn: vanilla inpainting_a} \\
&  \boldsymbol{a}_{i - 1}^{\text{known}}   \sim p_{i-1}^{\text{known}} = \mathcal{N}\left( {\sqrt{{\bar{\alpha}}_{t}}\boldsymbol{a}_{0}^{\text{known}},\left( {1 - \bar{\alpha}} \right)I} \right) \label{Eqn: inpt_known_a} \\
&  \boldsymbol{a}_{i - 1}^{\text{unknown}}   \sim q_{i-1}^{\text{unknown}} = \mathcal{N}\left( {\mu_{\theta}\left( {\boldsymbol{a}_{i}, O, i} \right),\Sigma_{\theta}\left( {\boldsymbol{a}_{i}, O, i} \right)} \right) \label{Eqn: inpt_unknown_a} 
\end{align} 
The diffusion policy is executed in a receding-horizon control fashion~\cite{chi2023diffusion}. As illustrated in Figure~\ref{Fig: inpainting}(a), at time $t$, the diffusion policy utilizes historical observations $O_t$ with length $T_o$ to generate a predicted multi-step action sequence $\mathbf{a}_t$ with length $T_p$, and executes the first $T_a$ actions. At time $t+T_a$, the policy enters the next cycle, repeating the observation and trajectory generation until the task is completed. 
During the generation of the action sequence, inpainting is applied by setting the known part as the keyframes, i.e., $\boldsymbol{a}_0^{\text{known}} = \boldsymbol{a}^{\text{key}} \cdot \boldsymbol{m}$. In this setup, the mask $\boldsymbol{m}$ for the first $T_o+T_a - 1$ steps is set to 0, and subsequent steps are set to 1. Thus,  $\boldsymbol{a}^{\text{key}}$ serves to condition the generation of  $\boldsymbol{a}^{\text{unknown}}$. 
\rev{When one task contains multiple keyframes, DISCO selects the first keyframe and sequentially switches to the following ones when the current keyframe is achieved (two consecutive actions remain almost unchanged).}
\rev{For grasp pose generation, where only a single action is executed, inpainting optimization is applied by fixing the 3D position from the keyframe and sampling the rotation component with the diffusion model, ensuring the final 6D grasp pose is both conditioned and feasible.}

\begin{algorithm}[t]
\caption{DISCO for Manipulation} \label{Alg: disco}
\textbf{Inputs: } Environment, language description $l$, observation $O_0$, VLM with prompts, diffusion policy model ($\mu_{\theta}$ and $\Sigma_{\theta}$)\;
Reasoning key points for task: $\boldsymbol{p}^{\text{key}} \gets  \text{VLM}(O_0, l)$\;
Generate action keyframe: $\boldsymbol{a}^{\text{key}} \gets \text{Env.}(\boldsymbol{p}^{\text{key}})$ \;
Formulate initial history observation by padding $O=[0, 0, \dots, O_0]$, and set $t=0$ \;
\While {task is \textbf{NOT} Done}
{
    $\boldsymbol{a}_N \sim \mathcal{N}(\boldsymbol{0}, \boldsymbol{I})$\;
    \For {$i$ in $\{N, \dots, 1\}$}
    {
    Perform reversed diffusion to calculate  $\mu_{\theta}\left( {\boldsymbol{a}_{i}, O, i} \right),\Sigma_{\theta}\left( {\boldsymbol{a}_{i}, O, i} \right)$ \;
    Perform inpainting optimization $\boldsymbol{a}_{i-1}$ in Eqn.~\eqref{Eqn: simple inpainting} and \eqref{Eqn: inpt_known_a}, given $\boldsymbol{a}_0^{\text{known}} = \boldsymbol{a}^{\text{key}}$. \;
    }
    Obtain $\boldsymbol{a}_0$, execute $\boldsymbol{a}_0[t:t+T_a-1]$ in Env. \;
    Update time step, $t \gets t+T_a$ \;
    Collect new observation $O=[O_{t-T_o+1}, \dots, O_t]$ \;
}
\end{algorithm}

\subsection{Inpainting Optimization for Out-of-Distribution Keyframes} \label{Subsec: inpainting opt}

Vanilla inpainting can produce unsuccessful robot behavior due to ``poor'' keyframes --- it forces the generation to comply with the  keyframe which leads to suboptimal $\boldsymbol{a}^{\text{unknown}}$. This results in the robot state moving towards regions that are not well supported by the training data and consequently, and the produced behavior fails to accomplish the task.

To address this issue, we propose a constrained inpainting optimization approach. Intuitively, we adjust the generation process to ensure that  when faced with poor keyframes, the generated action sequences remain within the data distribution. This approach helps sustain task performance even under the challenging conditions introduced by novel task descriptions. More formally, during the reserve step of the diffusion model, 
we replace Eqn.~\eqref{Eqn: vanilla inpainting_a} with,
\vspace{-1mm}
\begin{align}
\boldsymbol{a}_{i-1}  & = {\arg\min\limits_{\boldsymbol{a}_{i - 1}}{D^{\text{key}}\left( {\boldsymbol{m} \odot \boldsymbol{a}_{i - 1}^{\text{known}},  \boldsymbol{m} \odot \boldsymbol{a}_{i - 1}} \right)}} ,\label{Eqn: inpt opt obj} \\
& \text{s.t.} \ -\log\left( q_{i-1}^{\text{unknown}}(\boldsymbol{a}_{i-1}) \right) \leq \gamma_{i - 1} ,
\label{Eqn: inpt opt const}
\end{align}
where, $\boldsymbol{a}_{i-1}^{\text{known}}$ denotes the known part in Eqn.~\eqref{Eqn: inpt_known_a} and $q_{i-1}^{\text{unknown}}$ is density in Eqn.~\eqref{Eqn: inpt_unknown_a}. 
$D^{\text{key}}$ measures the distance between the generated $\boldsymbol{a}_{i-1}$ and the $\boldsymbol{a}_{i - 1}^{\text{known}}$ (keyframe $\boldsymbol{a}^{\text{key}}$). We design the distance function $D^{\text{key}}$ according to the policy and task environment. 
We constrain the negative log-likelihood $-\log\left( q_{i-1}^{\text{unknown}}(\boldsymbol{a}_{i-1}) \right)$ to be bounded by hyperparameter $\gamma_i$.

Figure~\ref{Fig: inpainting}(b) illustrates inpainting optimization: the left diagram shows that the keyframe is within a high-likelihood region of $p(\boldsymbol{a}_{i - 1}^{\text{known}})$, i.e., $-\log\left( q_{i-1}^{\text{unknown}}(\boldsymbol{a}_{i-1}) \right) \leq \gamma_{i - 1}$. In this scenario, inequality Eqn.~\eqref{Eqn: inpt opt const} is always satisfied, and thus the formula reduces to vanilla inpainting (Eqn.~\eqref{Eqn: vanilla inpainting_a}).
In the right diagram, since $-\log\left( q_{i-1}^{\text{unknown}}(\boldsymbol{a}_{i-1}) \right) \geq \gamma_{i - 1}$, constrained optimization via Eqn (\ref{Eqn: inpt opt const})) results in a generated action sequence that lies within the training data distribution but also minimizes the distance to the keyframe. We find this enables the policy to complete the task while maximally satisfying the language specification.

The distance function $D^{\text{key}}$ and $\gamma_i$ are application-dependent. In our experiments, we set $D^{\text{key}}$ as the L-2 norm distance function and $q_{i-1}^{\text{unknown}}$ is Gaussian distribution. As such, we can simplify general constrained inpainting to the convex optimization problem,
\begin{align} \label{Eqn: simple inpainting}
&  \quad \boldsymbol{a}_{i-1}=  \underset{\boldsymbol{a}_{i-1}}{\operatorname{argmin}} \sum_{k, \boldsymbol{m}^{[k]}=1}\left(\boldsymbol{a}_{i-1}^{[k]}-\boldsymbol{a}_{i-1}^{\text{known}, [k]}\right)^2,  \\
& \begin{aligned}
    &  \text{s.t.} \ \frac{1}{2}\left[\boldsymbol{a}_{i-1}- \mu_{\theta}\left( {\boldsymbol{a}_{i},), i} \right) \right]^T \cdot \Sigma_{\theta}^{-1}\left( {\boldsymbol{a}_{i},O,i} \right) \\
    &  \qquad \qquad \qquad \qquad \cdot \left[\boldsymbol{a}_{i-1}-  \mu_{\theta}\left( {\boldsymbol{a}_{i},O,i} \right) \right] \leq \gamma_{i-1}^{\prime},  
\end{aligned} \\
& \qquad  \gamma_i^{\prime}=\gamma_i-\frac{1}{2} \log \left[(2 \pi)^d \Sigma_{\theta}\left( {\boldsymbol{a}_{i},O,i} \right) \right],
\end{align}
where superscript $[k]$ denotes the $k_{\text{th}}$ element of the vector. $d$ denotes the dimension of the actions. $\gamma_i^{\prime}$ is the normalized constraint based on $\gamma_i$ and the covariance matrix (see the Appendix for detailed ablation study of $\gamma_i$). 

\section{Experiment} \label{Sec: experiment}

% ===============
\begin{figure*}
    \centering
    \subfloat[Simulation experimental environments.]{\includegraphics[width=.27\textwidth,valign=m]{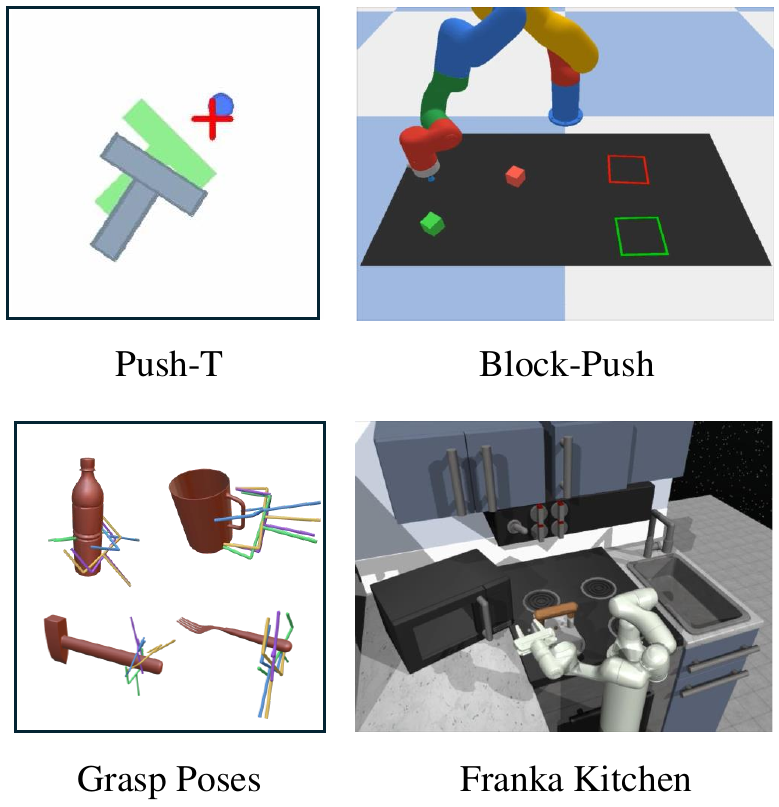}}
    \hfill
    \subfloat[Results of Simulation Experiments (success rate $\uparrow$)]
    {
    \resizebox{.7\textwidth}{!}{
    \renewcommand\arraystretch{1.3}
    % \begin{threeparttable}
    \begin{tabular}{ll|cc|cc|cccc}
\toprule
\multicolumn{2}{l|}{\multirow{2}{*}{}}                                      & \multirow{2}{*}{Push-T} & \multirow{2}{*}{Push-block} & \multicolumn{2}{c|}{Kitchen} & \multicolumn{4}{c}{Grasp Pose Generation} \\
\multicolumn{2}{l|}{}                                                       &                         &                             & Single        & Multi        & Mug    & Bottle   & Ham.   & Fork   \\ \hline
\multicolumn{1}{l|}{\multirow{4}{*}{Seen tasks $^1$}}   & No condition           & 0.54                    & 0.50                        & 0.12          & 0.03         & 0.47   & 0.31     & 0.40   & 0.48   \\
\multicolumn{1}{l|}{}                              & LangDiff~\cite{ha2023scaling} & \textbf{0.98}                    & \textbf{1.00}                        & \textbf{1.00}          & 0.31         & 0.91   & 0.84     & 0.78   & 0.96   \\
\multicolumn{1}{l|}{}                              & GoalDiff~\cite{reuss2023goal}  & 0.96                    & \textbf{1.00}                        & \textbf{1.00}          & 0.27         & 0.88   & \textbf{0.95}     & 0.70   & 0.93   \\
\multicolumn{1}{l|}{}                              & Vanilla inpt.  & 0.82                    & \textbf{1.00}                        & 0.71          & 0.35         & 0.90   & 0.88     & 0.85   & 0.96   \\
\multicolumn{1}{l|}{}                              &  DISCO(ours)     & 0.87                    & \textbf{1.00}                        & 0.71          & \textbf{0.42}         & \textbf{0.92}   & 0.92     & \textbf{0.97}   & \textbf{0.97}   \\ 

\hline
\multicolumn{1}{l|}{\multirow{4}{*}{Unseen tasks $^2$}}                              & LangDiff & 0.63                    & 0.63                        & -             & 0.06         & 0.02   & 0.04     & 0.02   & 0.24   \\
\multicolumn{1}{l|}{}                              & GoalDiff  & 0.83                    & 0.70                        & -             & 0.06         & 0.00   & 0.08     & 0.00   & 0.23   \\
\multicolumn{1}{l|}{}                              & Vanilla inpt.  & 0.89                    & 0.70                        & -             & 0.06         & 0.02   & 0.30     & 0.58   & 0.55   \\
\multicolumn{1}{l|}{}                              &  DISCO (ours)     & \textbf{0.94}                    & \textbf{0.75}                       & -             & \textbf{0.31}        & \textbf{0.65}   & \textbf{0.51}     & \textbf{0.80}   & \textbf{0.87}   \\ \toprule
\end{tabular}

% \end{threeparttable}
} % end resize box
    } % end subflot
    
\caption{
Experiments in diffusion policy-related environments.
\textbf{(a)}: Environments include
Push-T~\cite{florence2022implicit}, Block-push~\cite{shafiullah2022behavior}, Franka kitchen~\cite{sharma2018multiple}, and Grasp Pose generation~\cite{urain2023se} (details in website). 
\textbf{(b)}: The policies were only trained in the seen tasks and zero-shot transferred to unseen tasks. Success rates are calculated over 50 trials. Overall, we observed DISCO, which combines VLM-generated keyframes with constrained inpainting, achieved robust performance across seen and unseen tasks.
} % end caption
\label{Fig: main exp results}
\vspace{-5mm}
\end{figure*}

In this section, we discuss experiments designed to evaluate two key hypotheses related to DISCO:
\begin{enumerate}
    \item[\textbf{H1}:] Using VLM-generated keyframes enables effective language-guided completion of tasks.
    \item[\textbf{H2}:] Constrained inpainting improves task completion compared to standard inpainting, especially on novel task descriptions.  
\end{enumerate}

We conducted three sets of experiments. In the following, we first describe diffusion policy-related environments for trajectory and grasp pose generation (Sec.~\ref{Subsec: sim env}). 
Then, we tested DISCO on a large-scale language-conditioned manipulation benchmark, CALVIN (Sec.~\ref{Subsec: Calvin}). 
Finally, we evaluated zero-shot transfer from simulation to a real robot setup (Sec.~\ref{subec: real robot}).  
\rev{Experiment details are on \href{https://sites.google.com/view/disco2025/}{website}.}

\vspace{-0.7mm}
\subsection{Diffusion Policy Environments} \label{Subsec: sim env}
\vspace{-0.7mm}

\textbf{Experiment and Environment Setup}.
As shown in Figure~\ref{Fig: main exp results}, we adapted four simulation environments for language-conditioned manipulation tasks. We generated corresponding language descriptions and subgoals for the action sequences in the dataset, and categorized tasks into ``seen tasks'' and ``unseen tasks''. Seen tasks are those with demonstrations and corresponding language descriptions in the dataset.  Unseen tasks had \emph{new} language descriptions, e.g., tasks specified in a different way or using synonyms, or manipulating objects in a different order than in the dataset. 
In the 2D environments (push-T and block-push), DISCO generates key points in the 2D plane, which are converted to the positional keyframes. In the 3D grasp pose and Kitchen environments, DISCO observes the scene from three views to generate key points and 3D positional keyframes as in Section~\ref{Subsec: keyframe gen}.

\rev{To evaluate H1, we compared DISCO against an \textit{unconditional diffusion policy}~\cite{chi2023diffusion, urain2023se, urain2022se3dif} and alternative methods for language-conditioning, i.e., classifier-based diffusion models with a fine-tuned \textit{language-conditioned} diffusion policy (LangDiff)~\cite{ha2023scaling} or a \textit{goal-conditioned} diffusion policy (GoalDiff)
~\cite{reuss2023goal}
. We also conducted comparative tests to keyframe adherence using standard or  \textit{vanilla inpainting} (Vanilla inpt.)~\cite{lugmayr2022repaint} to evaluate {H2}.}

\textbf{Results}.
Overall, the results support both H1 and H2. As shown in Table~\ref{Fig: main exp results}(b), DISCO achieved comparable performance to both language and goal conditioned diffusion policies on \emph{seen tasks} --- the diffusion baselines performed better in tasks like push-T, push-block, and Franka kitchen (single task), but on the Franka kitchen composite/multi tasks and grasp pose generation, vanilla inpainting and DISCO achieved higher success rates. On \emph{unseen tasks}, the success rates of the DISCO are significantly higher than the other baseline methods. One potential reason is that the inpainting methods leverage the keyframes generated by VLMs, which have been trained on large amounts of text data and were able to produce reasonable keyframes in many circumstances. Performance is further enhanced when using constrained inpainting across all the tasks, supporting H2 and indicating the importance of generating action sequences within high-likelihood regions of the diffusion policy.
\rev{DISCO’s ability to handle unseen objects depends on both the VLM’s keyframe reasoning and the coverage of the downstream policy. When the policy is trained only on specific objects, generalization to new categories is limited. However, with unconditional diffusion models trained on diverse objects~\cite{urain2022se3dif}, DISCO can generate effective grasp poses even for unseen categories. DISCO's two-stage setup is flexible in that it enables us to swap in diffusion models dependent on downstream context.
}

\vspace{-0.7mm}
\subsection{CALVIN Benchmark}\label{Subsec: Calvin}
\vspace{-0.7mm}

\begin{figure}
    \centering
    \subfloat[Calvin Benchmark. ABC$\rightarrow D$ zero-shot transfer tasks.]{\includegraphics[width=.9\columnwidth,valign=m]{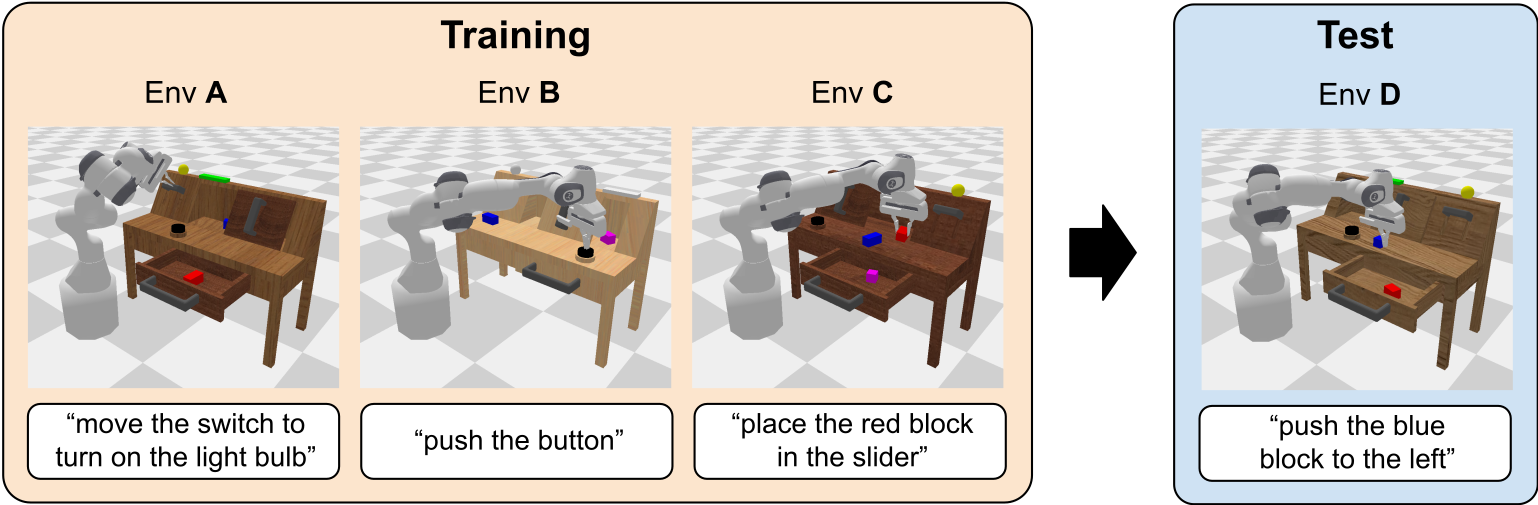}}
    % \hfill
    
    \subfloat[Results in Calvin benchmark (success rate $\times 100 \uparrow$)]
    {
    \resizebox{.95\columnwidth}{!}{
    \renewcommand\arraystretch{1.3}
\begin{tabular}{l|lllll|l}
\toprule
& \multicolumn{5}{c|}{Task completed in a row} &   \multirow{2}{*}{\makecell[c]{Avg. \\ Len}}       \\
& 1       & 2      & 3      & 4      & 5      & \\ 
\hline
RT-1~\cite{brohan2022rt}              & 53.3    & 22.2   & 9.4    & 3.8    & 1.3    & 0.90     \\
RoboFlamingo~\cite{li2023vision}      & 82.4    & 61.9   & 46.6   & 33.1   & 23.5   & 2.48     \\
SuSIE~\cite{black2023zero}             & 87.0    & 69.0   & 49.0   & 38.0   & 26.0   & 2.69     \\
3D Diffuser Actor~\cite{ke20243d} & 92.2    & 78.7   & 63.9   & 51.2   & 41.2   & 3.27     \\
Vanilla inpt. & 92.9     & 78.6   & 64.9    & 51.4    & 41.9    & 3.30
    \\
DISCO (ours)       & \textbf{94.7}     & \textbf{82.9 }    & \textbf{71.0}   & \textbf{58.8}     & \textbf{49.4}
     & \textbf{3.57}      \\ \toprule
\end{tabular}

} % end resize box
} % end subflot
\caption{
Experiments in Calvin Benchmark~\cite{mees2022calvin}. Results of DISCO were calculated in $3 \times 1000$ trials.
} % end caption
\label{Fig: calvin exp results}
\vspace{-8mm}
\end{figure}

\textbf{Experiment and Environment Setup}. The CALVIN benchmark~\cite{mees2022calvin} (Fig.~\ref{Fig: calvin exp results}(a)) was built for language-conditioned policy learning in long-horizon robot manipulation tasks. 
Each instruction chain comprises five language instructions that need to be executed sequentially. 
Our experiments involved the \textit{zero-shot generalization setup}, where models are trained in environments ABC and tested in D. We implemented DISCO with ChatGPT4 as keyframe generator and 3D Diffuser Actor~\cite{ke20243d} as the low-level diffusion policy. DISCO observes and marks key points in the top and front views and synthesis target end-effector position as the keyframe to guide the policy. We compared DISCO with SOTA language-conditioned policies RT-1~\cite{brohan2022rt}, RoboFlamingo~\cite{li2023vision}, and SuSIE~\cite{black2023zero} and DISCO with vanilla inpainting method. Following prior work~\cite{mees2022calvin}, we report the success rate and the average length of completed sequential tasks.

\textbf{Results.} 
As shown in Table~\ref{Fig: calvin exp results}(b), DISCO achieved the best performance on the zero-shot transfer (ABC$\rightarrow$D) tasks, with an average length of 3.57, improving upon the SOTA baseline policy. Vanilla inpainting did not improve performance --- this was primarily due to the inaccurate keyframes that caused the suboptimal action generation. 
We observed that DISCO mainly improved success rates in two types of scenarios: those requiring (i) additional observational information and (ii) common-sense understanding of the language instructions. For example, the task \emph{"Place the grasped block into the slider"} requires the robot to first grasp and lift one block and then place it into the slider. However, CALVIN does not provide precise observations of the slider (only whether it ``left'' or ``right'') and the policy has to infer its position and whether it is open/closed from the training dataset. DISCO exploits the image processing capabilities of VLMs to infer the state/position of the slider and generate sufficiently useful keyframes. This led to an increase of task success from $70\%$ to $90\%$. For the task \emph{``Push the red/blue/pink block to the left/right side''}, the language instruction was not in the dataset and requires the policy to understand that it is required to push the block in a particular direction. The baseline policy struggled with this task, achieving only a $50\%$ success rate, but by using the common-sense understanding of VLMs, DISCO improved the success rate to $85\%$.

\revcolorbegin
\vspace{-1mm}
\subsection{Additional Analysis}
\vspace{-0.7mm}

\textbf{Keyframe quality analysis.}
We investigate how the clarity of task instructions influences the quality of keyframe generation—specifically, whether the resulting keyframes are reasonable and satisfy the intended task conditions. For seen tasks (e.g., ``pick up the mug by the handle''), instructions are explicit and conditions are well defined. In contrast, unseen tasks present moderate ambiguity (e.g., ``Give me the bottle'') or high ambiguity (e.g., ``Hand over the hammer to me''), where object references or conditions are indirect or missing. We assess keyframe quality using two metrics: (1) the improvement in DISCO’s success rate over the vanilla inpainting method, and (2) the normalized deviation of generated actions from the provided keyframes ($D^{\text{key}}$ in Eqn.~(12)). 
As summarized in Table~\ref{Tab: Keyframe Quality Analysis}, most tasks involve clear instructions that yield high-quality keyframes, with minimal improvement or deviation required. As ambiguity increases, DISCO demonstrates larger gains over vanilla inpainting, and the deviation from the diffusion policy distribution increases accordingly. In highly ambiguous cases, keyframes become nearly out-of-distribution, making the inpainting optimization essential for correcting and refining the guidance. This analysis highlights how instruction ambiguity directly affects keyframe quality and underscores the importance of constrained inpainting in handling suboptimal guidance.

\begin{table}[t]
\centering
\caption{\rev{Keyframe Quality Analysis}}
\label{Tab: Keyframe Quality Analysis}
\renewcommand\arraystretch{1.4}
\resizebox{.75\columnwidth}{!}{
\begin{tabular}{lcccc}
\hline
 \makecell[c]{Instruction \\ clearness} & \makecell[c]{Task \\ portion} & \makecell[c]{Success rate \\ improvement $\downarrow$} & \makecell[c]{Keyframe \\ deviation $\downarrow$} \\
\hline
Clear & 0.44 & 0.04 & 0.07 \\
Moderate & 0.40 & 0.12 & 0.25 \\
Ambiguous & 0.16 & 0.22 & 0.41 \\
\hline
\end{tabular}
}
\vspace{-6mm}
\end{table}

\textbf{Comparison with Visual-Language-Action (VLA) Models.}
VLA models, such as OpenVLA~\cite{kim2024openvla}, $\pi_0$~\cite{black2024pi_0}, 
% Octo~\cite{team2024octo}, 
and DeeR-VLA~\cite{yue2024deer}, have recently gained popularity as an alternative approach for language-guided robotics, relying on fine-tuning large vision-language models to generate robot actions end-to-end via auto-regressive outputs. While these models provide a powerful solution, DISCO adopts a different strategy by decoupling high-level reasoning and low-level control: VLMs are used to generate interpretable keyframes, which then guide a diffusion policy through inpainting optimization. We compare DISCO with the recent DeeR-VLA model in the CALVIN zero-shot setup (ABC$\rightarrow$D), where DISCO achieves a higher average task completion length (3.57) than DeeR-VLA (2.90). This result highlights the advantage of VLM-guided diffusion policies for compositional and open-vocabulary manipulation tasks.

\revcolorend

\vspace{-0.7mm}
\subsection{Transfer to Real Robot Grasping} \label{subec: real robot}

\begin{figure}
    \centering
    \subfloat[Real-robot experiments setup]{\includegraphics[width=.90\columnwidth,valign=m]{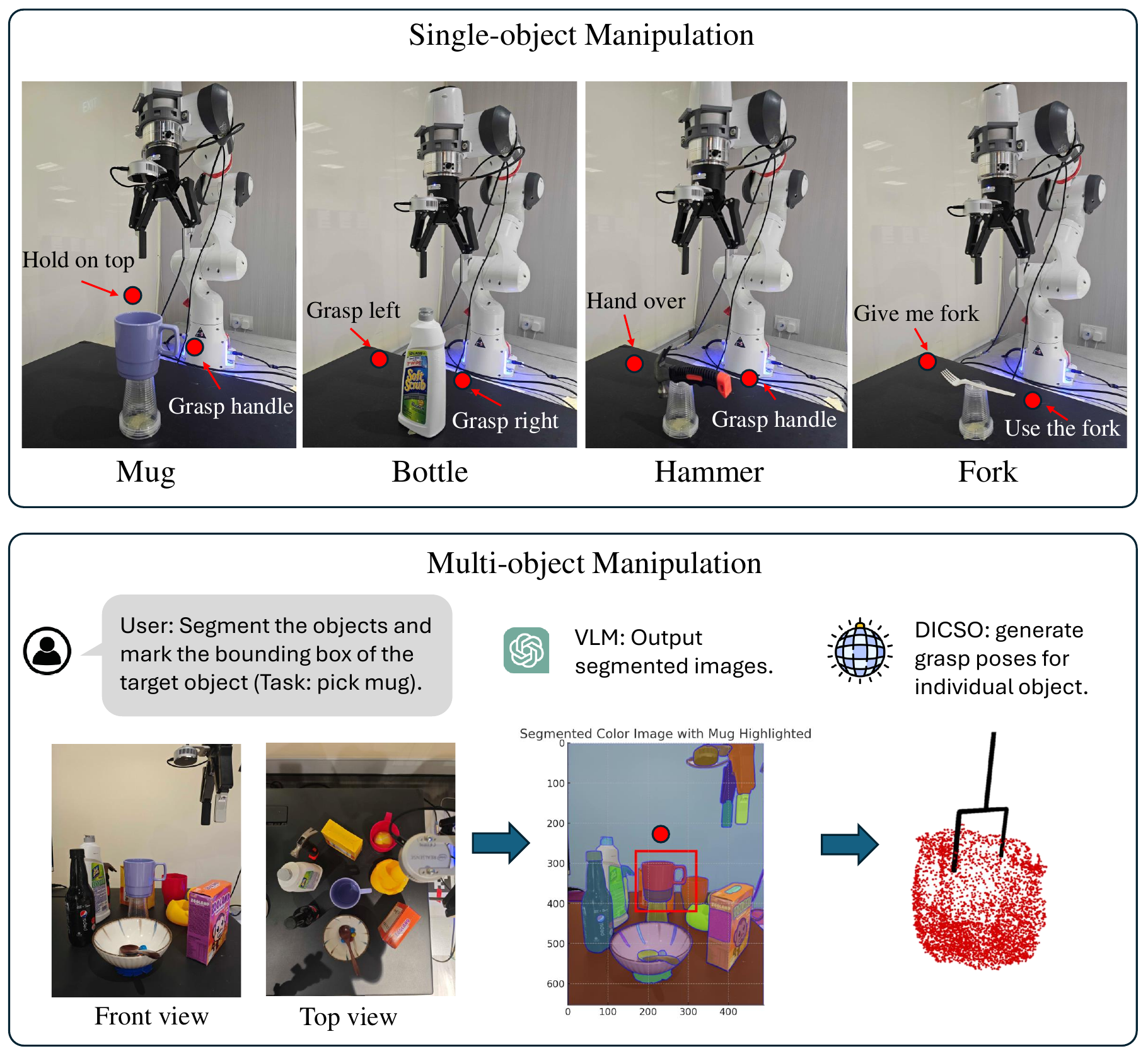}}
    % \hfill
    
    \subfloat[Results in Real-robot Experiments (success rate $\uparrow$)]
    {
    \resizebox{.8\columnwidth}{!}{
    \renewcommand\arraystretch{1.3}
\resizebox{.9\columnwidth}{!}{
\begin{tabular}{l|cc|cc}
\hline
              & \multicolumn{2}{c|}{Single-object task} & \multicolumn{2}{c}{Multi-object task} \\
              & Seen              & Unseen         & Seen          & Unseen       \\ \hline
GoalDiff~\cite{reuss2023goal}    & 0.10              & 0.0           & 0.05          & 0.0         \\
Vanilla inpt. & 0.55              & 0.15          & 0.30          & 0.05         \\
DISCO (ours)  & \textbf{0.85}     & \textbf{0.65} & \textbf{0.65} & \textbf{0.25} \\ \hline
\end{tabular}
} % end resize box
}
} % end subflot
\caption{
Real-robot experiments. We transferred DISCO from simulation to real-robot scenarios with multiple objects. 
In the single-object tasks, DISCO generated language-conditioned grasp poses for one object, while in multi-object tasks, we first segmented the target object from the cluster, and then applied DISCO. 
Each object was tested with 20 trials.
} % end caption
\label{Fig: real exp results}
\vspace{-7mm}
\end{figure}

\textbf{Experiment and Environment Setup}. We transferred a diffusion-based grasp pose generation model trained in simulation to a real-world Franka Emika robot (Figure ~\ref{Fig: real exp results}). We applied DISCO to generate 3D keyframes given real images and point clouds, then generated conditional grasp poses via constrained inpainting. We used a \emph{single} grasp diffusion model~\cite{urain2022se3dif} trained in simulation to generalize across all object categories without any parameter fine-tuning. Example language instructions for seen tasks are ``give me the mug from top'', and ``grasp the hammer handle'', while unseen tasks include ``lift the bottle" and "hand over the fork''.

\textbf{Results.} 
Overall, the results in Table.~\ref{Fig: real exp results}(b) support both H1 and H2.
In the \emph{single-object} manipulation experiments, 
The goal-conditioned methods achieved only a $10\%$ success rate in seen tasks and $0\%$ in unseen tasks. On seen tasks, vanilla inpainting and DISCO (inpainting optimization) methods achieved success rates of $55\%$ to $85\%$, respectively. In unseen tasks where vanilla inpainting faltered ($15\%$), DISCO maintained a $65\%$ success rate.
In the \emph{multi-object} manipulation experiment, DISCO still achieved fairly good grasping in seen tasks with success rate $65\%$. The tasks with ``grasping from the side'' instruction only achieved $40\%$ successes, but when asked to pick objects ``from the top'' success rates rose to $80\%$ as this avoided collisions with other objects. We admit this limitation that the keyframes generated by DISCO did not consider the potential collisions when the end-effector moves to it. Finally, DISCO could also understand the unseen instruction in cluttered scenarios and achieve certain successful grasping (25 \%).

\vspace{-2mm}
\section{Conclusion, Discussion, and Future Work} \label{Sec: conclusion}
\vspace{-1mm}

In this paper, we propose DISCO, a zero-shot, open-vocabulary framework for robot manipulation using language descriptions.
By leveraging VLMs, our method guides diffusion policies through actionable keyframes and addresses critical out-of-distribution issues with an inpainting optimization technique. This approach is able to match (and sometimes surpass) traditional fine-tuned methods in both simulated and real-world settings. 
We believe DISCO represents a key step towards effective language-conditioned robot behavior. While it performs well in our experiments, there are areas for improvement. At present, DISCO does not account for occlusions during reasoning, 
which could be addressed by incorporating object pose information, applying joint constraints, or using failure data to replan and gather more information. 
DISCO is also heavily reliant on the underlying diffusion policy, which could be further improved by novel flow-matching models.
\rev{Another limitation is that when DISCO executes a sequence of sub-tasks, keyframe switching is based on an empirical thresholding rule, which can occasionally lead to backtracking or suboptimal transitions.
}
We are actively working to build upon DISCO to address these limitations.

%%%%%%%%%%%%%%%%%%%%%%%%%%%%%%%%%%%%%%%%%%%%%%%%%%%%%%%%%%%%%%%%%%%%%%%%%%%%%%%%

%\section*{Acknowledgements}

% \clearpage
% 
\balance
\bibliographystyle{IEEEtran}
\bibliography{references}

\clearpage
\balance

\appendix
% \appendices

\subsection{Simulation Environments} \label{Appendix: sim env}

In the simulation environments, we designed specific settings for "Push-T", "Block-push", "Franka Kitchen" and "grasp pose" to test our diffusion policy and inpainting methods under various task conditions.

\textbf{Push-T}: A blue round end-effector pushes a gray T-block towards a fixed green T-shaped target. Successful task completion is based on the alignment between the T-block and the target. The end-effector starts from the target's top-right, while the T-block starts from the bottom-left, each with random positional and rotational deviations. The task descriptions involve detouring the block from specified sides (left, right, top, down), with unseen conditions being TOP and DOWN due to the lack of demonstration trajectories from these directions, which may cause the policy to fail by colliding with the block.

\textbf{Block-Push}: This involves pushing red and green blocks into designated target squares, structured in two phases—moving one block followed by the other. Random deviations in initial positions introduce unpredictability in successful trajectories. Demonstrations exist for prioritizing either block, forming the basis for task descriptions that dictate the order of operations. Unseen tasks include non-existent color blocks to test the algorithm's response to erroneous instructions and open-vocabulary tasks that challenge the system's interpretative flexibility.

\begin{table*}[b]
\centering
\renewcommand\arraystretch{1.3}
\caption{Language Task Descriptions for Different Environments}
\label{tab:task_descriptions}
\resizebox{.99\textwidth}{!}{
\begin{tabular}{|l|m{6cm}|m{6cm}|}
\hline
\textbf{Environment} & \makecell[c]{\textbf{Seen Tasks}} & \makecell[c]{\textbf{Unseen Tasks}} \\ \hline
Push-T & \textbullet \ Push the block to the target region and detour from LEFT side. \newline \textbullet \ Push the block to the target region and detour from RIGHT side. & \textbullet \ Push the block to the target region and detour from TOP side. \newline
\textbullet \ Push the block to the target region and detour from DOWN side.\\ \hline
Block-push & \textbullet \ Push the RED block to the target, then the GREEN block to its target. \newline \textbullet \ Push the GREEN block to the target, then the RED block to its target. & \textbullet \ Push the YELLOW block to the target, then the BLUE block to its target. \newline  \textbullet \ Push two blocks to targets in any sequence. \\ \hline
\makecell[l]{Franka Kitchen \\ single-task} & \textbullet \ Complete task \textbf{A}. (A is one of the sub-tasks in the Kitchen environment.) & - \\ \hline
\makecell[l]{Franka Kitchen \\ multli-task} & \textbullet \ Complete task \textbf{A}, then task \textbf{B}. \newline (Demonstrations have trajectories with A to B order.) & \textbullet \ Complete task \textbf{C}, then task \textbf{D}. \newline (Demonstrations do not have trajectories in C to D order.) \\ \hline
\makecell[l]{Grasp Pose: \\ Mug} & 
\textbullet \ Grasp the mug RIM (top). \newline \textbullet \ Grasp the mug HANDLE. & 
\textbullet \ Give me the mug. \newline \textbullet \ Grasp the mug BOTTOM. \\ \hline
\makecell[l]{Grasp Pose: \\ Bottle} & 
\textbullet \ Grasp the bottle on LEFT side. \newline \textbullet \ Grasp the bottle on RIGHT side. & 
\textbullet \ LIFT the bottle. \newline \textbullet \ Grasp the bottle BOTTOM. \\ \hline
\makecell[l]{Grasp Pose: \\ Hammer} & 
\textbullet \ Grasp the hammer HANDLE. \newline \textbullet \ Grasp the hammer HEAD. & 
\textbullet \ USE the hammer. \newline \textbullet \ HAND OVER the hammer. \\ \hline
\makecell[l]{Grasp Pose: \\ Fork} & 
\textbullet \ Grasp the fork HANDLE. \newline \textbullet \ Grasp the fork HEAD. & 
\textbullet \ PICK up the fork. \newline \textbullet \ HAND OVER the fork. \\ \hline

\end{tabular}
}
\end{table*}

\textbf{Franka Kitchen}: Comprising seven sub-tasks grouped into three levels based on their locations, this environment tests the robot arm's ability to perform sequential tasks as specified by the task descriptions. All single tasks are seen during demonstrations, making them familiar, while multiple tasks involve unseen sequences, especially when changing from high to mid or low levels, or vice versa, challenging the policy's adaptability.

\textbf{Grasp Pose Generation}: This environment focuses on the precise task of generating grasp poses for four different objects: mugs, bottles, hammers, and forks. Each object has associated tasks, divided into seen and unseen categories, where the seen tasks involve specific, demonstrated grasping actions, and the unseen tasks introduce new, potentially open-vocabulary or challenging conditions.

The environment is designed to test the diffusion policy's ability to adapt its output (a 6D grasp pose) based on the conditions provided by the task descriptions. Keyframes are generated from the positions in 3D space, which guide the policy in generating actions conditioned by the specific grasp requirements. The key point here is the direct application of conditions to the action generation through inpainting, which does not involve temporal sequencing but rather focuses on spatial accuracy and relevance.

Each environment utilizes a diffusion policy where actions are determined by the 2D or 3D positions of keyframes generated via vision-language models. This setup tests both the precision of task execution based on seen instructions and the flexibility of the system under unseen or open-vocabulary conditions. The task descriptions for each environment are listed in Table~\ref{tab:task_descriptions}.

\clearpage

\begin{figure*}[t]
\centering
\begin{minipage}{0.97\textwidth}

\begin{tcolorbox}[colback=gray!10, colframe=gray!50, boxrule=0.5pt, arc=2mm, left=1mm, right=1mm, top=1mm, bottom=1mm, title={Prompts for Keypoints Generation.}]

\textbf{Inputs:} 

[task description] is: "\textit{Pick the mug from the table to the cupboard.}" 

[multi-view images] are: "front view", "left view" and "top view". (Load images with pre-defined sequences.)

\begin{center}
\includegraphics[width=0.6\columnwidth]{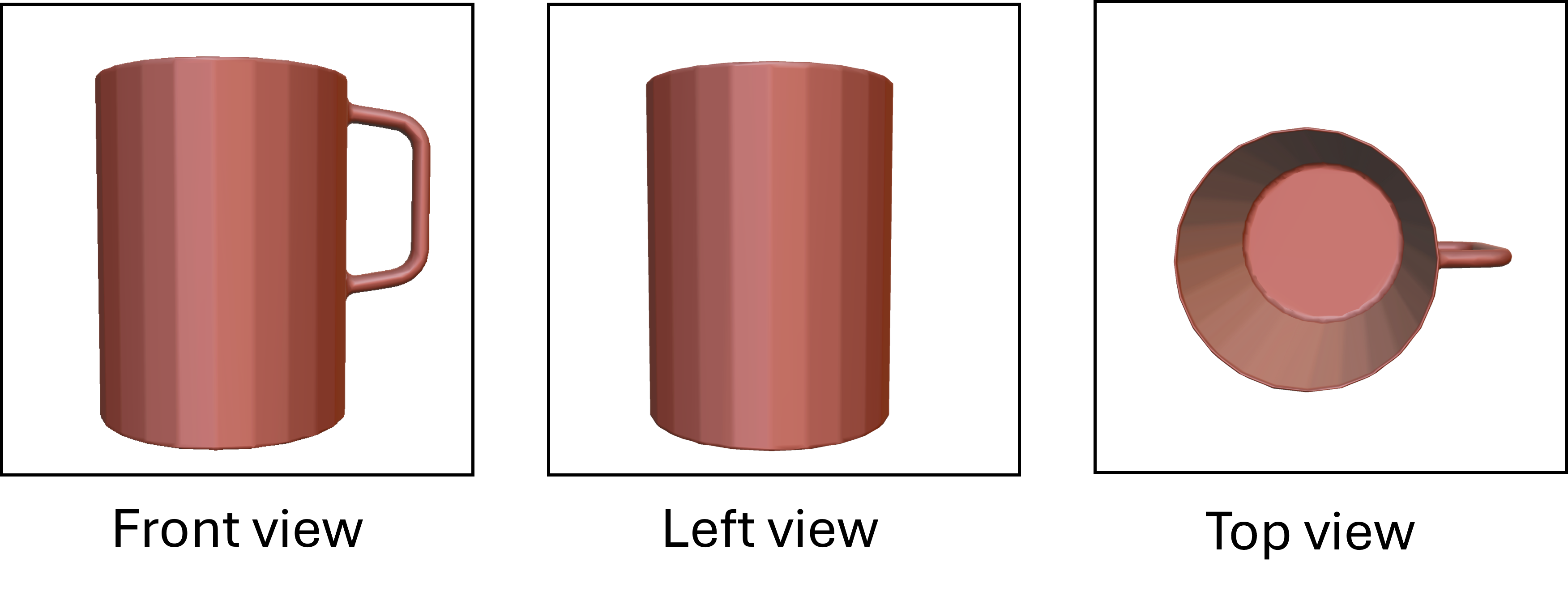}
\end{center}

\vspace{-2mm}

\textbf{System Prompts:}
\begin{itemize}
    \item You need to finish a robot manipulation task using a gripper hand.
   \item I will give you a [task description] and [multi-view images] to finish the task.
   \item The [task description] includes the main target and conditions, and you should consider both requirements.
   \item Please build a coordinate in the [multi-view images], in the front view, the bottom left is 0, and the upper right is 1. The x-axis points to the right, the y-axis points up, and the z-axis points outside. The coordinate from the other view will change according to the front view.
\end{itemize}

\textbf{Action Prompts:}
\begin{enumerate}
    \item According to [task description], and content in the [multi-view images], write the [key steps] to finish the task.
    \item Read the [task description] and [multi-view images], and build the coordinate.
    \item According to the [key steps], find the corresponding [key point] in the images. Mark the point in the front view and write the position (x,y,z).
\end{enumerate}

\vspace{3mm}

\textbf{Outputs:}

\textbf{1. Key Steps Reasoning}

\begin{itemize}
    \item \textbf{Perception and Localization:}
    \begin{itemize}
        \item Detect the mug position and orientation from multi-view images.
        \item Compute 3D coordinates of graspable area (mainly around the mug body or handle).
    \end{itemize}
    
    \item \textbf{Planning Grasp:}
    \begin{itemize}
        \item Determine a stable grasping pose:
        \begin{itemize}
            \item For a gripper hand, the side grasp (on the mug body) is preferred due to handle presence.
            \item Avoid grasping the handle directly to ensure stability.
            \item Approach the mug from the front or slightly above.
        \end{itemize}
    \end{itemize}

    \item \textbf{Grasp Execution:}
    \begin{itemize}
        \item Move the gripper to the target grasp position.
        \item Close the gripper securely around the mug.
    \end{itemize}

    \item \textbf{Transfer Movement:}
    \begin{itemize}
        \item Lift the mug vertically to avoid collision with the table.
        \item Move horizontally towards the cupboard.
    \end{itemize}

    \item \textbf{Placement:}
    \begin{itemize}
        \item Lower the mug into the cupboard placement area.
        \item Release the gripper to place the mug safely.
    \end{itemize}

    \item \textbf{Retreat:}
    \begin{itemize}
        \item Move the gripper away to avoid collision after placement.
    \end{itemize}
\end{itemize}

\end{tcolorbox}
\end{minipage}
\end{figure*}

\clearpage
 
% ==========================================

\begin{figure*}[t]
\centering
\begin{minipage}{0.97\textwidth}

\begin{tcolorbox}[colback=gray!10, colframe=gray!50, boxrule=0.5pt, arc=2mm, left=1mm, right=1mm, top=1mm, bottom=1mm, title={Continue}]

\textbf{2. Position Estimation from Multi-view Images}

\begin{itemize}
    \item \textbf{Front View (Image 1):} The handle is not visible, which suggests the handle is likely oriented sideways from this perspective (either left or right).
    
    \item \textbf{Left View (Image 3):} The handle is clearly visible on the right-hand side of the mug. In this view:
    \begin{itemize}
        \item Horizontal axis: $z$ direction.
        \item Vertical axis: $y$ direction.
        \item The handle center is roughly at:
        \begin{itemize}
            \item $z \approx 0.75$
            \item $y \approx 0.5$
        \end{itemize}
    \end{itemize}
    
    \item \textbf{Top View (Image 2):} The handle is seen extending to the right of the mug. In this view:
    \begin{itemize}
        \item Horizontal axis: $x$ direction.
        \item Vertical axis: $z$ direction.
        \item The handle center is approximately at:
        \begin{itemize}
            \item $x \approx 0.75$
            \item $z \approx 0.5$
        \end{itemize}
    \end{itemize}
\end{itemize}

% ======================

\textbf{3. Key Point Selection Based on Task \& Images}

\begin{itemize}
    \item \textbf{From Front View (Reference Frame):} The handle is not directly visible in the front view.
    
    \item Based on multi-view alignment, the handle lies to the right side (positive $x$-direction), roughly at:
    \begin{itemize}
        \item \textbf{Front View Mapping:}
        \begin{itemize}
            \item Horizontal position ($x$): around $0.75$
            \item Vertical position ($y$): around $0.5$
        \end{itemize}
    \end{itemize}
    
    \item Thus, in the front view, the handle keypoint is approximately at: $(x=0.75,\, y=0.5)$
\end{itemize}

\begin{center}
\includegraphics[width=0.6\columnwidth]{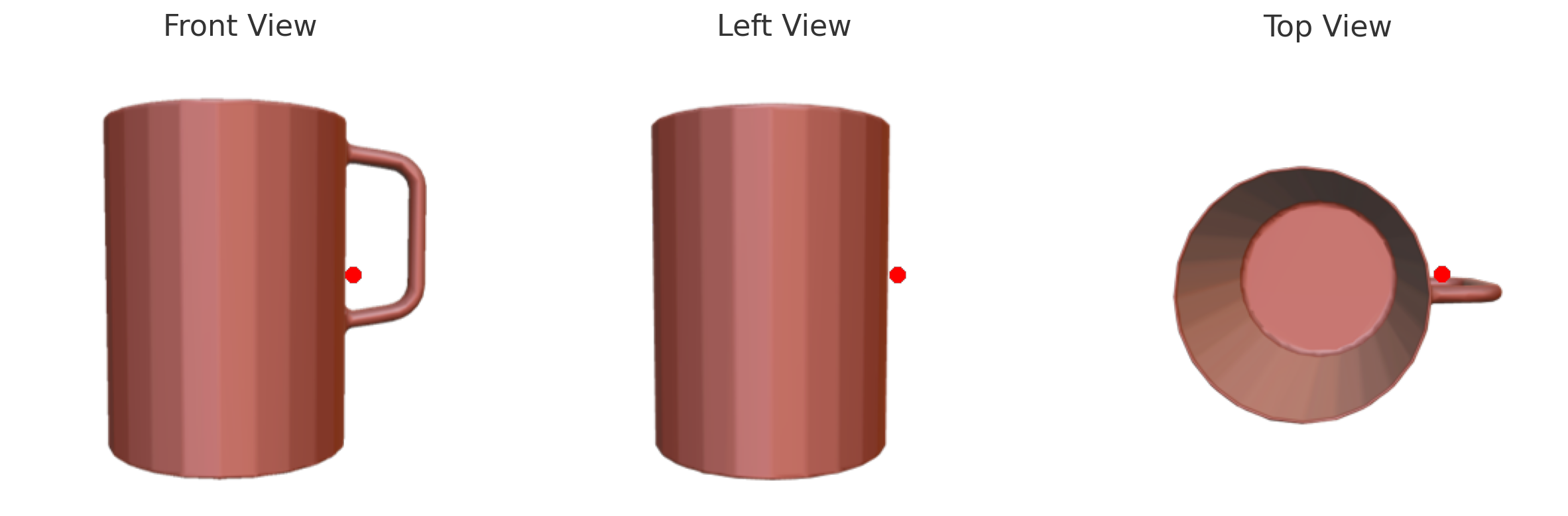}
\end{center}

Here are the annotated images:

\begin{itemize}
    \item \textbf{Front View:} keypoint marked at $(x=0.75,\, y=0.5)$.
    \item \textbf{Left View:} keypoint mapped at $(z=0.75,\, y=0.5)$.
    \item \textbf{Top View:} keypoint mapped at $(x=0.75,\, z=0.5)$.
\end{itemize}

This reflects the estimated center of the handle for a stable grasp.

\end{tcolorbox}
\end{minipage}
\end{figure*}

\revcolorbegin

\subsection{Keyframe Generation and Conditioning Details}

In Section~\ref{Subsec: keyframe gen} and Fig.~\ref{Fig: keyframe}, we show the keyframe generation processes. We provide the details for keyframe generation and conditioning. In the first two steps of reasoning the key steps to finish the task and marking the keypoints in the multi-view images, we utilized the VLMs (ChatGPT4o) with prompts as above. With the reasoning results, we can obtain keypoints and project them into the 2D or 3D spaces with calibrated mapping. 

Since the observation and action spaces vary in different environments, DISCO used pre-collected data to calibrate the image-to-position mapping~\cite{li2023manipllm}. In the 2D environments (push-T and push-block), each point directly maps a $(x,y)$ position in the plane. In the grasp pose generation environments, images are clipped and calibrated to $(x,y,z)$ in 3D space for the front, left, and top views. In the Franka Kitchen and Calvin environments, the image observation is from the third-person view, we use the demonstrations with images and actions to calibrate the end-effector position in the image pixel space. 

In addition, the action space of the push-block and Calvin environments is velocity (delta pose). The keyframes are correspondingly calculated by the delta of the ideal and last end-effector poses. Also, for the Franka Kitchen environment, the action space is the velocity of joint angles, so DISCO adopted the inverse kinematic model of the Franka robot to convert the end-effector pose to the joint values. In short, the velocity controlling rules are obtained similarly to model-based control with first-order differential equations of dynamics and inverse kinematic models.

In closed-loop trajectory planning environments, DISCO determines when to switch the keyframe condition by continuously monitoring the robot's state during execution. Specifically, at each control step, the system compares the 6D pose of the end-effector at the current and previous time steps to the keyframe. If the positional and rotational distance falls below a small predefined threshold, DISCO considers the keyframe achieved and transitions to the next one (or no condition). This threshold-based approach is simple, robust, and effective in practice, ensuring that the policy only proceeds once the intended keyframe is sufficiently satisfied.

\subsection{Experimental Details in Calvin Environments}

\begin{figure}[t]
    \centering
    \includegraphics[width=0.97\columnwidth]{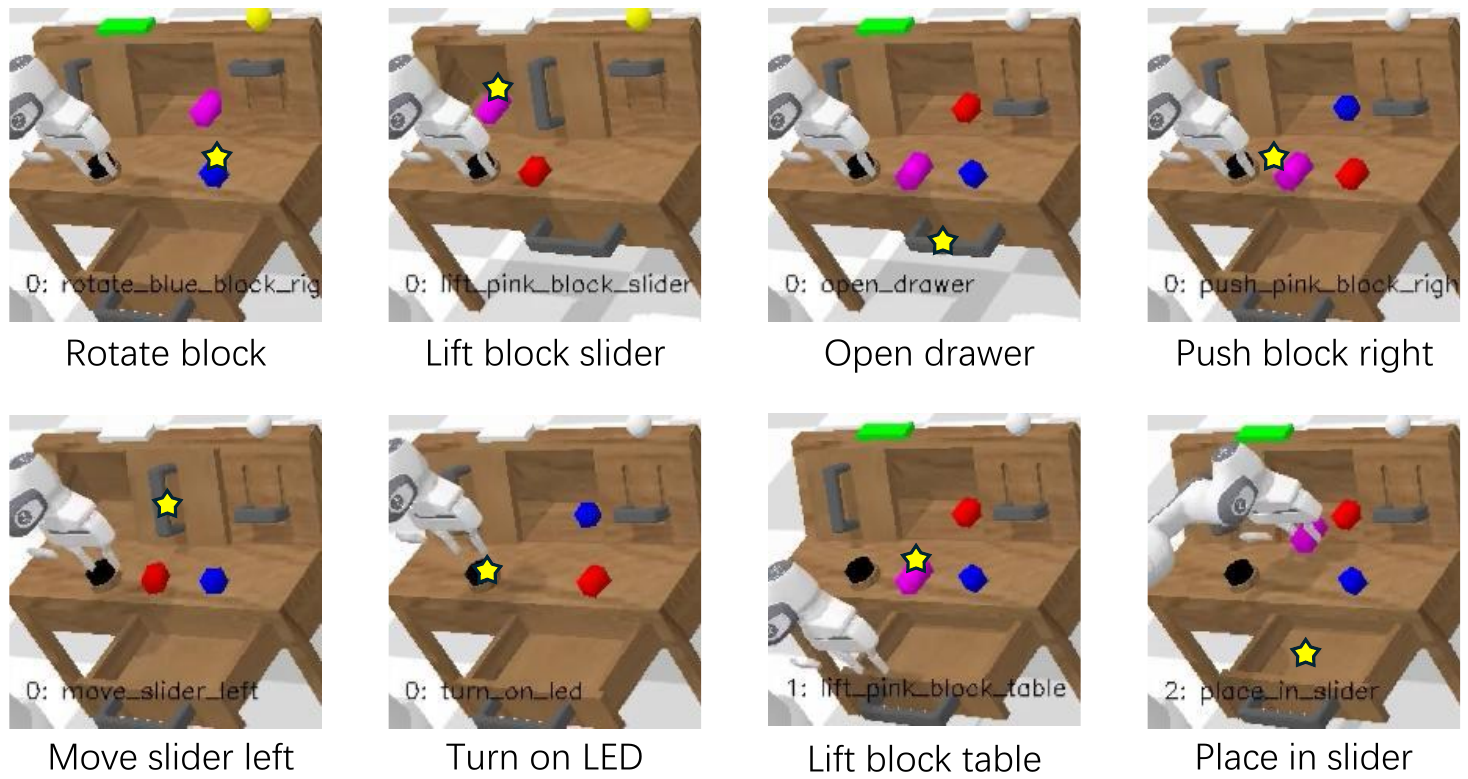}
    \caption{Illustration of keypoints $\bigstar$ in the Calvin environment. The pixels in the third-person views are projected into 3D position space~\cite{li2023manipllm}.}
    \label{Fig: keyframe calvin}
\end{figure}

\begin{table}[t]
\centering
\caption{Task-level evaluation in Calvin environment}
\label{Tab:calvin_eval}
\resizebox{.97\columnwidth}{!}{
\renewcommand\arraystretch{1.3}
\begin{tabular}{lccc}
\hline
\makecell[l]{\textbf{Breakdown} \\ \textbf{sub-tasks}} & 
\makecell[c]{Baseline policy\\success rate~\cite{ke20243d}} & 
\makecell[c]{DISCO\\success rate} & 
\makecell[c]{Success rate\\improvement} \\
\hline
Rotate block L/R          & 0.84 & 0.85 & 0.01 \\
Push block L/R            & 0.56 & 0.85 & \textbf{0.29} \\
Move slider L/R           & 0.91 & 0.94 & 0.03 \\
Open/close drawer         & 0.92 & 0.93 & 0.01 \\
Lift block table          & 0.91 & 0.94 & 0.03 \\
Lift block slider         & 0.89 & 0.95 & 0.06 \\
Lift block drawer         & 0.88 & 0.91 & 0.03 \\
Place into slider/drawer  & 0.73 & 0.89 & \textbf{0.16} \\
Push into drawer          & 0.88 & 0.91 & 0.03 \\
Stack blocks              & 0.85 & 0.93 & 0.08 \\
Unstack blocks            & 0.84 & 0.91 & 0.07 \\
Turn on/off light bulb    & 0.90 & 0.95 & 0.05 \\
Turn on/off LED           & 0.88 & 0.91 & 0.03 \\
\hline
\end{tabular}
}
\end{table}

In the CALVIN environment~\cite{mees2022calvin}, we evaluated zero-shot language generalization on the ABC$\rightarrow$D task suite. To implement DISCO, task instructions and images are first provided to the VLM for keyframe generation. As illustrated in Fig.~\ref{Fig: keyframe calvin}, the VLM interprets both the language instruction and visual scene to predict relevant grasping positions, marked as key points (stars in the figure). These are then mapped to 3D coordinates using position calibration~\cite{li2023manipllm}, and serve as end-effector keyframes for guiding trajectory generation.

Section~\ref{Subsec: Calvin} reports overall task completion results, while Table~\ref{Tab:calvin_eval} details the breakdown of success rates across sub-tasks. We find that for most tasks, both the baseline policy~\cite{ke20243d} and DISCO achieve high success rates ($>0.85$). However, three sub-tasks exhibit relatively lower baseline performance: ``Push block L/R,'' ``Rotate block L/R,'' and ``Place into slider/drawer.'' For ``Rotate block L/R,'' the main challenge is accurately perceiving block orientation from images, which limits the policy’s performance and cannot be effectively resolved by DISCO. In ``Push block L/R,'' the policy struggles to interpret the directional language, whereas DISCO utilizes the VLM to resolve ``left'' or ``right'' in 3D space, yielding a substantial improvement ($+0.29$). Similarly, in ``Place into slider/drawer'' tasks, the VLM can directly localize the slider or drawer in the image, and keyframe-based guidance with DISCO leads to a significant success rate gain ($+0.16$). These findings highlight the value of VLM-guided keyframes in tasks where spatial reasoning from ambiguous language or image cues is required.

\revcolorend

% ========================================
\subsection{Real-robot Experiments} \label{Appendix: real}

\begin{figure}[t]
    \centering
    \includegraphics[width=0.6\columnwidth]{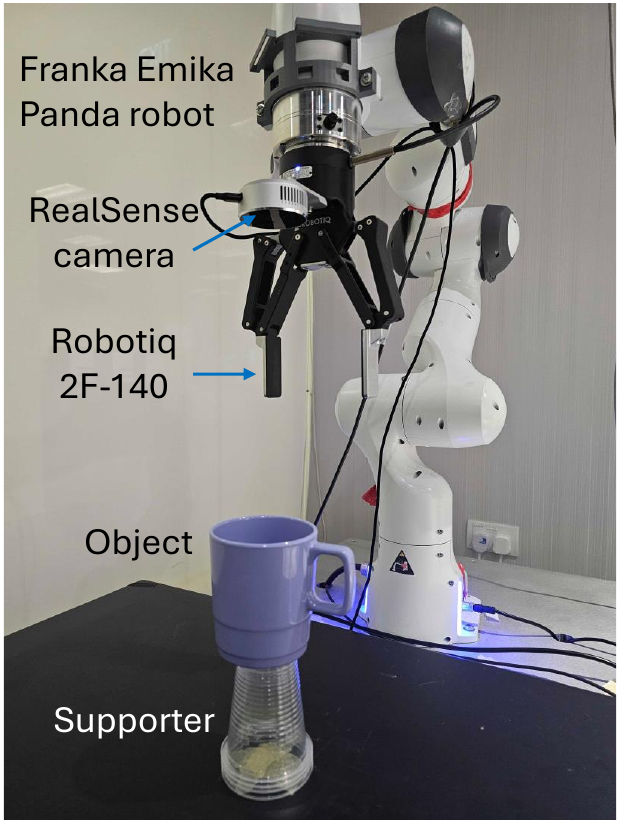}
    \caption{Real-robot experiment setup.}
    \label{Fig: real app}
\end{figure}

The real robot experiments adopted the Franka Emika robot to grasp and manipulate different objects according to task descriptions. The RealSense camera observed the multi-view RGB-D images and generated point clouds as observations. The end-effector was the Robotiq 2F-140 gripper. A supporter was placed to lift the objects, which enlarged the workspace of the robot arm. We filtered out the supporters in the point cloud. For each object, we transferred the trained model in simulation to the real robot and sampled 10 grasps for seen and unseen tasks (Table~\ref{tab:task_descriptions}). 

One big challenge in the real-world experiment was the partial observation of the point cloud. Unlike point clouds sampled from simulated meshes, the camera-generated point clouds had an uneven density over different regions. For example, the mug handle has a sparser point cloud density than the rim. This problem led to a strong Sim2Real gap and failures of fine-tuned conditional models.

% ==================================

\subsection{Ablation Study of Inpainting Optimization} \label{Appendix: ablation}

\begin{figure*}[t]
    \centering
    \includegraphics[width=0.8\textwidth]{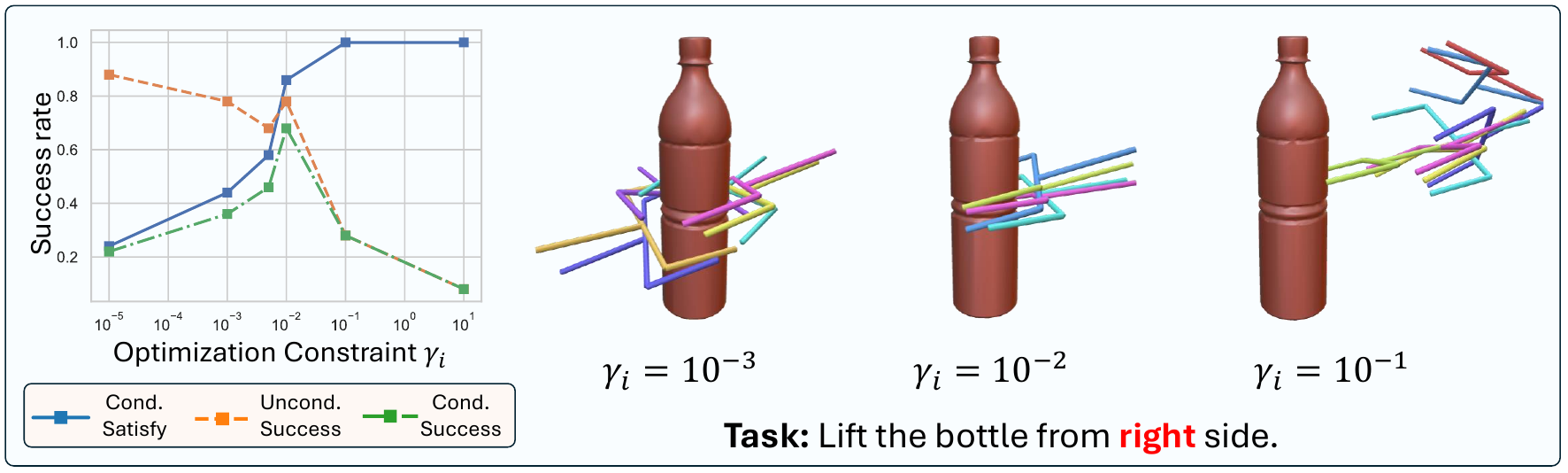}
    \caption{\textbf{Ablation study} of inpainting optimization in grasp pose environment (keyframes omitted in the figures). As the constraint $\gamma_i$ increases, the generated actions increasingly satisfy the specified conditions. However, excessively high values compromise the overall success of the task. A carefully chosen $\gamma_i$ balances condition satisfaction against task success.}
    \label{Fig: gamma ablation}
\end{figure*}

We study the influence of the optimization constraint $\gamma_i$ in Eqn.~\eqref{Eqn: simple inpainting} on the performance of inpainting optimization and provide a general method to tune this hyperparameter. In Figure~\ref{Fig: gamma ablation}, we applied DISCO on a grasp pose generation task of 'Lift the bottle from the right side'. DISCO generated a keyframe on the right side, but not closely attached to the bottle. Our goal was to use the keyframe to guide the diffusion policy and generate successful grasps that satisfy the condition. 

Then we varied the constraint $\gamma_i$ from $10^{-5}$ to $10^1$. 
When $\gamma_i$ is relatively small $10^{-3}$, the generated actions align closely with the unconditional demonstration, achieving successful grasping, but they often fail to meet the specified conditions. As $\gamma_i$ increases to $10^{-1}$, inpainting optimization compels the actions to satisfy these conditions more rigorously, but this adherence to the keyframes can substantially lower the success rate.
Therefore, a moderate $\gamma_i=10^{-2}$ can strike an optimal balance between condition fulfillment and alignment with the demonstration distribution, thereby maximizing the conditional success rate.
For instance, in three cases: 
\begin{itemize}
    \item $\gamma_i = 10^{-3}$: Smaller $\gamma_i$ keeps poses close to the demonstration distribution, resulting in various grasp positions but often failing to meet the 'right side' requirement.
    \item $\gamma_i = 10^{-1}$: Larger $\gamma$ aligns poses closer to the keyframe condition, but many grasps fail to lift the bottle.
    \item $\gamma_i = 10^{-2}$: Balances between the keyframe and the demonstration distribution, achieving successful grasps that meet the condition.
\end{itemize}
Therefore, the $\gamma_i$ can be generally selected by scanning its value and optimizing it to maximize the conditional success rate. 
In practice, we used $\gamma_i=10^{-2}$ for end-effector position keyframe, $\gamma_i=10^{-3}$ for end-effector velocity, and $\gamma_i=3^{-4}$ for joint values.

\balance

\subsection{Networks, Datasets and Training Details} \label{Appendix: training}

For simulation experiments, we adapted the program from~\cite{chi2023diffusion} for push-T, push-block and Franka kitchen environments; and used the~\cite{urain2023se} for grasp pose generation environments. 
For diffusion policy networks that predict action sequences, we adopted the transformer-based backbone for state-based environments. For inpainting optimization, we formulated the convex optimization problem and utilized CVXPY to obtain the optimal solutions. 
We built the language-conditioned classifier network with text as input, followed by tokenization and encoding~\cite{ha2023scaling}. For the goal-conditioned classifier network, we augmented the network inputs with the normalized goal state~\cite{reuss2023goal}. Finally, we trained the classifier guidance networks to control the generation of diffusion models~\cite{dhariwal2021diffusion}. 

The training datasets were adapted from \cite{chi2023diffusion} and \cite{urain2023se}. We filtered and labeled the demonstrations with task descriptions for each environment. Note that in our experiments, we only used demonstrations that corresponded to pre-defined task descriptions and neglected other trajectories. In addition, we marked the terminal state of each trajectory as the goal state for goal-conditioned networks. For continuous action environments, we used batch size = 256, and for the grasp pose (single-action) environment, batch size = 2.
We trained networks in our local personal computer, that the CPU model is Intel(R) Core(TM) i9-14900KF and the GPU model is RTX 4090. During the training process, the average GPU memory usage is 12GB. The average training time for the diffusion model was 8 $\sim$ 10 hours.

\end{document}